\documentclass[10pt,twocolumn,letterpaper]{article}

\usepackage{wacv}              

\usepackage{graphicx}
\usepackage{amsmath}
\usepackage{amssymb}
\usepackage{booktabs}

\usepackage{multirow}
\usepackage{graphicx}
\usepackage{float}

\usepackage[pagebackref,breaklinks,colorlinks]{hyperref}

\usepackage[capitalize]{cleveref}
\crefname{section}{Sec.}{Secs.}
\Crefname{section}{Section}{Sections}
\Crefname{table}{Table}{Tables}
\crefname{table}{Tab.}{Tabs.}

\def\wacvPaperID{1014} 
\def\confName{WACV}
\def\confYear{2025}

\begin{document}

\title{Generalizable Single-view Object Pose Estimation by Two-side Generating and Matching}

\author{Yujing Sun\footnotemark[1]\\
The University of Hong Kong\\
China\\
{\tt\small yjsun@cs.hku.hk}
\and
Caiyi Sun\footnotemark[1]\\
Beijing Institute of Technology\\
China\\
{\tt\small scy639@outlook.com}
\and
Yuan Liu\footnotemark[1]\\
The University of Hong Kong\\
China\\
{\tt\small yuanly@connect.hku.hk}
\and
Yuexin Ma\\
Shanghai Technology University\\
China\\
{\tt\small mayuexin@shanghaitech.edu.cn}
\and
Siu Ming Yiu\\
The University of Hong Kong\\
China\\
{\tt\small smyiu@cs.hku.hk}
}

\maketitle
 
\renewcommand{\thefootnote}{\fnsymbol{footnote}} 
\footnotetext[1]{These authors contributed equally to this work.} 

\begin{abstract}
In this paper, we present a novel generalizable object pose estimation method to determine the object pose using only one RGB image. Unlike traditional approaches that rely on instance-level object pose estimation and necessitate extensive training data, our method offers generalization to unseen objects without extensive training, operates with a single reference image of the object, and eliminates the need for 3D object models or multiple views of the object. These characteristics are achieved by utilizing a diffusion model to generate novel-view images and conducting a two-sided matching on these generated images.
Quantitative experiments demonstrate the superiority of our method over existing pose estimation techniques across both synthetic and real-world datasets. Remarkably, our approach maintains strong performance even in scenarios with significant viewpoint changes, highlighting its robustness and versatility in challenging conditions. 
The code will be released at \url{https://github.com/scy639/Gen2SM}.
\end{abstract}


\section{Introduction}
\label{sec:intro}

\begin{figure}[thb] 
\centering
  \includegraphics[width=.48 \textwidth]{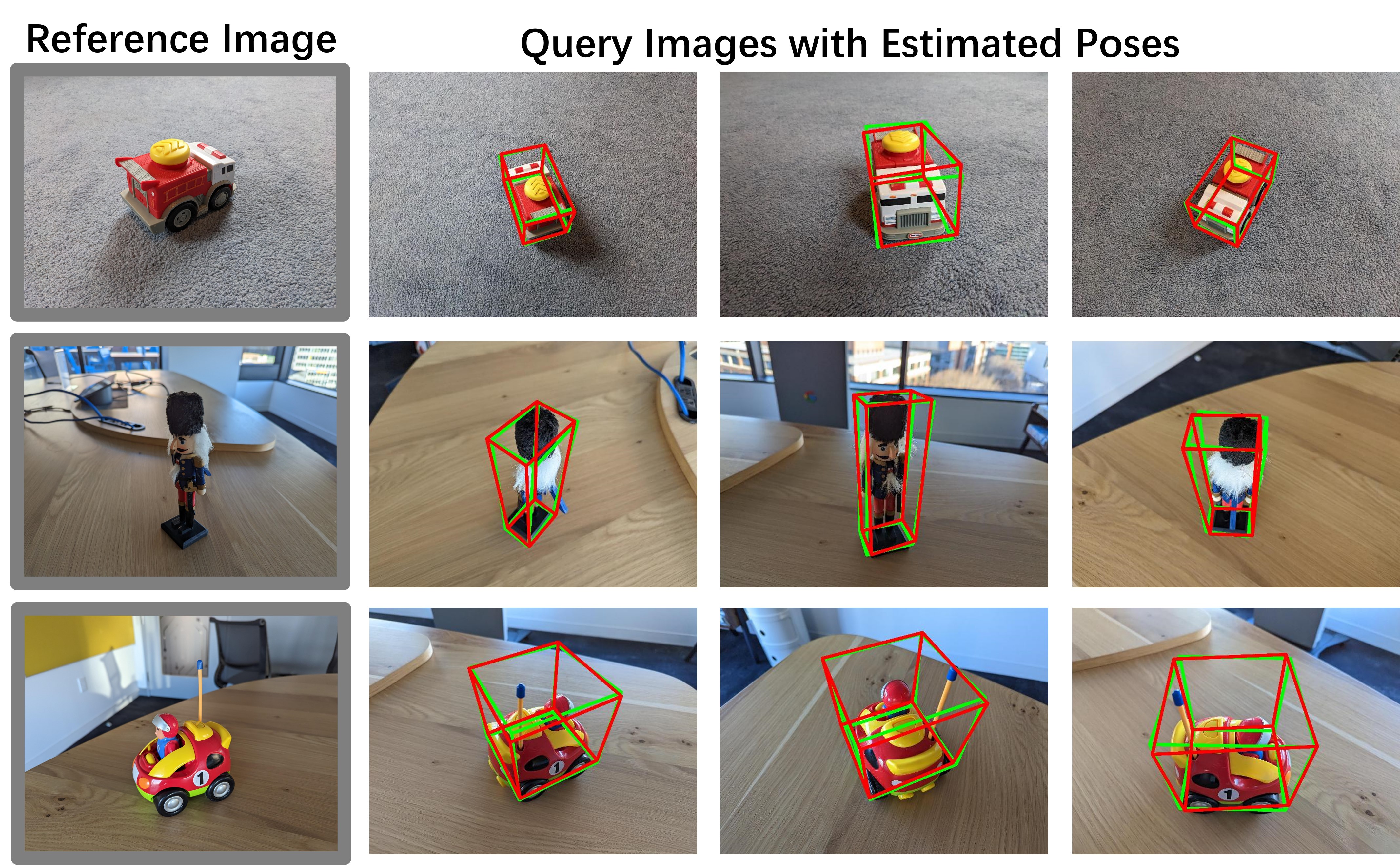}
  \caption{ 
Our method excels at accurately estimating object poses with a single reference image. Importantly, it maintains its accuracy even when faced with query images exhibiting substantial viewpoint changes from the reference image. The object poses estimated by our method offer significant value for a variety of practical applications such as 3D reconstruction.
}
  \label{fig:teaser}
\end{figure}

Recognizing object poses is an essential task in computer vision.
Although humans can easily recognize two different viewpoints of the same object, even when the object has never been seen before, such a seemingly straightforward skill presents a significant challenge for artificial intelligence. 
Early research in pose estimation was limited to estimating poses within the same category~\cite{kanazawa2019learning,kocabas2020vibe,ma2022virtual,usman2022metapose,wen2022disp6d},  
and relied on feature correspondence~\cite{liu2010sift,choy2016universal,sarlin2020superglue,sun2021loftr}, which perform poorly on new objects with large viewpoint changes because of the limited availability of effective features. 
In recent years, advancements in deep learning have led to increased focus on category-agnostic pose estimation~\cite{li2017deepim,zhao2022fusing,liu2022gen6d,sun2022onepose,he2022oneposeplusplus,10350449_PoseMatcher,zhao2023locposenet} and resulted in continuous accuracy improvement. 

Nevertheless, although these category-agnostic methods~\cite{li2017deepim,zhao2022fusing,liu2022gen6d,sun2022onepose,he2022oneposeplusplus,10350449_PoseMatcher,zhao2023locposenet} demonstrate effectiveness in predicting unseen object poses in some cases, they still suffer from two major challenges. 
The first is the limited generalization ability. 
These methods are usually trained on the ShapeNet~\cite{chang2015shapenet} or CO3D~\cite{co3D2021} datasets, which contain limited categories and data volume.
In comparison, the proficiency of humans in inferring huge viewpoint changes between images arises from the vast amount of data accumulated since childhood. 
Unlike the extensive exposure to information experienced by humans, the training data adopted by these methods is strongly biased to specific capturing patterns and object categories. 
The second challenge is that to generalize to unseen objects, these methods often require a renderable CAD model of this new object or densely captured multiview images of the object. However, reconstructing a renderable CAD model is costly, and capturing multiple images of the same object is tedious and complex.

\begin{figure}[thb] \centering
    \includegraphics[width=0.40\textwidth]{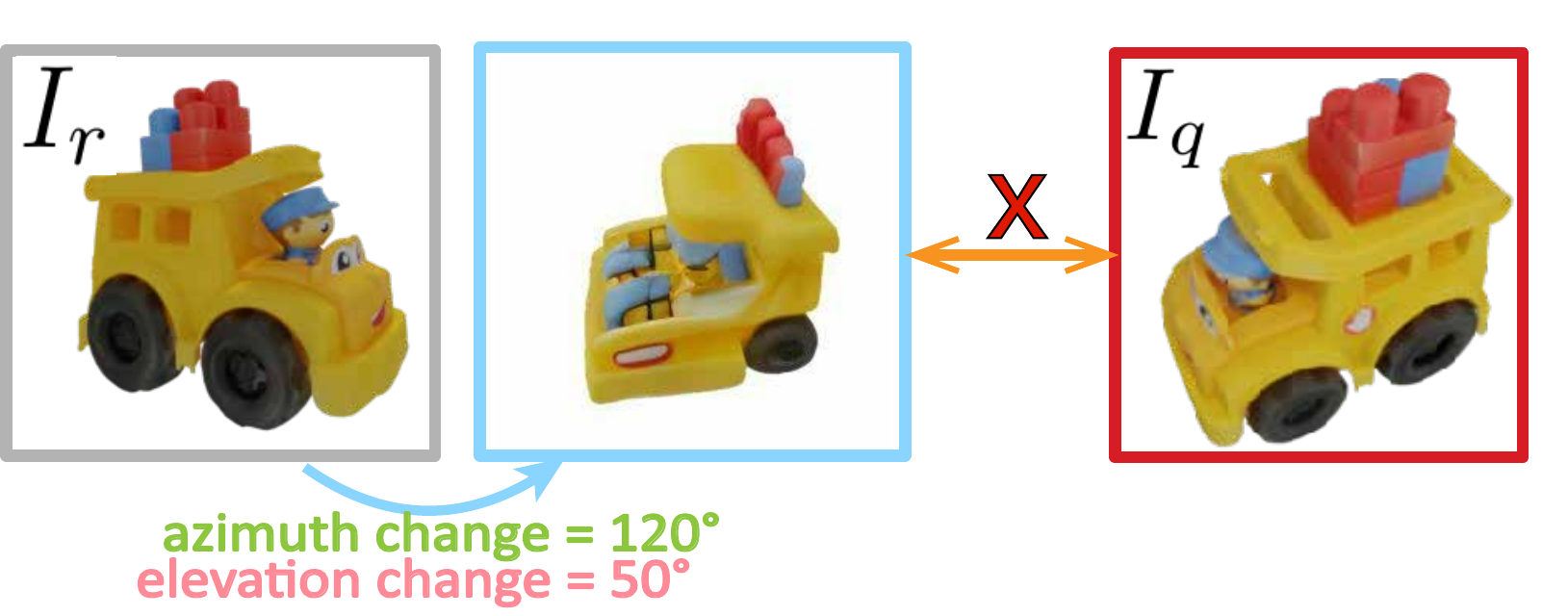}
    \makebox[0.235\textwidth]{(a)
    } 
    
    \includegraphics[width=0.48\textwidth]{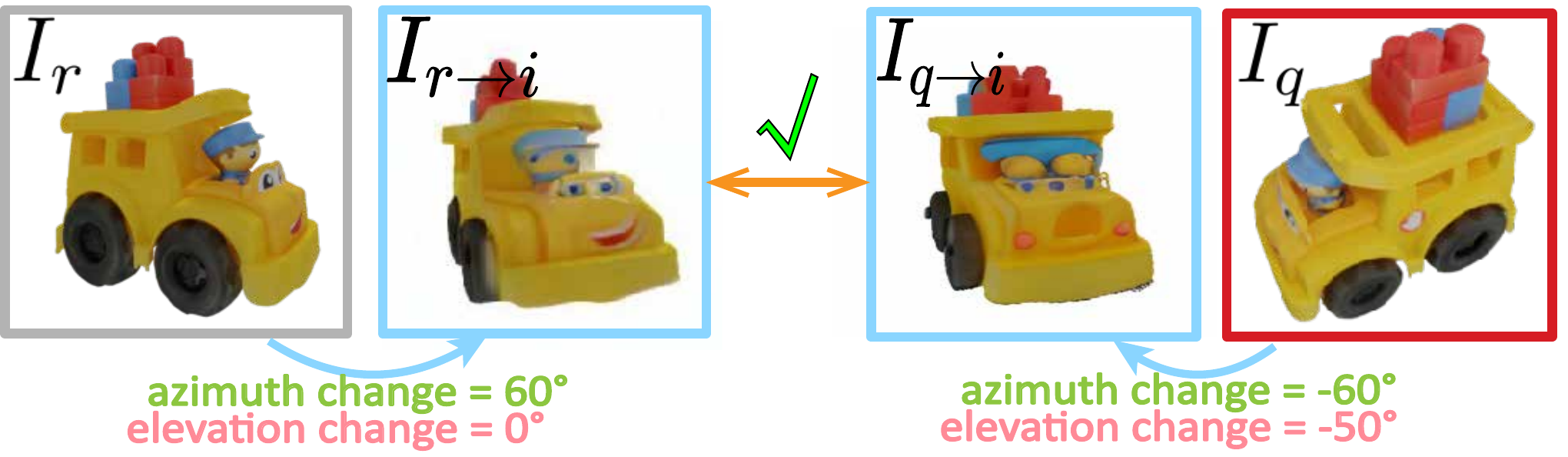}
    \makebox[0.235\textwidth]{(b)
    }
    \caption{
    (a) Utilizing Zero123~\cite{liu2023zero123} to directly generate an image from the viewpoint of the query image $I_q$ based on the reference image $I_r$. Due to the significant viewpoint change, the generated image exhibits low quality, hindering accurate matching with the query image. (b) Instead, we leverage Zero123~\cite{liu2023zero123} to generate images from intermediate viewpoints between the query image $I_q$ and the reference image $I_r$. These generated images at intermediate viewpoints show better alignment, facilitating the estimation of an accurate relative pose between $I_r$ and $I_q$.
    } \label{fig:good_and_bad_match}
\end{figure}

Some recent methods~\cite{e2vg,idpose2023} utilize diffusion models~\cite{liu2023zero123} to estimate relative camera poses showing promising generalization ability and only require images of the object.
E2VG~\cite{e2vg} and IDPose~\cite{idpose2023} leverage Zero123~\cite{liu2023zero123} trained on vast amounts of 3D objects, which significantly enhances generalization compared to previous approaches. 
E2VG~\cite{e2vg} employs a diffusion-based generative model Zero123~\cite{liu2023zero123} to produce a series of novel RGB views conditioned on a reference image, which is then matched with the query image.
IDPose~\cite{idpose2023} solves the problem by inverting the denoising diffusion process and minimizing the difference between the generated images and the input images to solve for the relative pose.
However, when there is a large viewpoint change between two input images, indicating a lack of shared regions, the generation quality of Zero123 degenerates severely, as shown in Fig.~\ref{fig:good_and_bad_match}, leading to difficulties in accurately matching the query image with the appropriate generated view. 
Both E2VG~\cite{e2vg} and IDPose~\cite{idpose2023} encounter this bottleneck.

To address this problem, we introduce a two-side matching module for object pose estimation.
Since the direct generation of results under significant viewpoint changes is not good enough for pose matching, we resort to adopting an incremental approach, that uses multiple small viewpoint changes to approximate a large viewpoint change. As a simple example shown in Fig.~\ref{fig:good_and_bad_match}, if reference image $I_r$ and the query image $I_q$ have a $120^{\circ}$ viewpoint difference in azimuth, directly estimating their relative pose can be very challenging. But this task can be simplified by using an image at an intermediate viewpoint as shown in Fig.~\ref{fig:good_and_bad_match} (b). Here, we generate an image at the intermediate viewpoint $I_{r\to i}$ (rotating $60^{\circ}$ from $I_r$) and $I_{q\to i}$ (rotated $60^{\circ}$ from $I_q$). Now the problem becomes to match between $I_{r\to i}$ and $I_{q \to i}$, which is much easier than matching between $I_r$ and $I_q$ directly. 
In our implementation, we use $N$ intermediate viewpoints and we call this matching scheme a two-side matching scheme. 
In comparison with E2VG~\cite{e2vg} and IDPose~\cite{idpose2023}, our two-side matching approach greatly reduces the challenges in generation and thus results in a significant performance improvement. 

Extensive experiments are conducted to verify our effectiveness. Convincing results are shown in Fig.~\ref{fig:teaser}. Remarkably, we outperform baseline methods by a large margin on both real dataset NAVI~\cite{jampani2023navi} and synthetic dataset GSO~\cite{gso2022google}, demonstrating the strong capability in estimating unseen object poses with large viewpoint changes. Furthermore, we showcase the integration of our method into an AR application for rendering new objects, as depicted in the supplementary material.
The contribution of this paper is summarized as follows:
\begin{itemize}
    \item We present an innovative pose estimation algorithm taking advantage of the diffusion model.
    \item Our approach demonstrates significant improvement in generalization capability. Without training, we outperform SOTA methods on both synthetic and real datasets.   
    \item The proposed two-side matching strategy overcomes the bottleneck in previous diffusion generative model-based approaches and demonstrates a dramatically improved performance. 
\end{itemize}



\section{Related Works}

\subsection{Generative Diffusion Models}
The application of neural networks in image generation tasks has led to the widespread use of Image Diffusion Models~\cite{ho2020denoising,rombach2022high,mou2023t2i,zhang2023adding,liu2023zero123,liu2023syncdreamer,long2023wonder3d}. These models aim to predict and remove noise from noisy images through many steps, generating clear, high-quality images.
To improve efficiency, researchers have focused on latent diffusion models~\cite{rombach2022high,liu2023zero123}. Instead of directly denoising the image in pixel space, the latent diffusion models concentrate on denoising the latent representation of the image.
Recent advancements in diffusion models~\cite{mou2023t2i,zhang2023adding} have allowed for conditioning the models on additional inputs. By incorporating these additional inputs, the models can generate images that align with specific requirements and constraints.
An exemplary latent diffusion model is Stable Diffusion~\cite{rombach2022high}, a text-to-image generative mode trained on Internet-scale image-text pairs data.
More recently, Zero-1-to-3 \cite{liu2023zero123}, SyncDreamer \cite{liu2023syncdreamer} and Wonder3D~\cite{long2023wonder3d} fine-tuned Stable Diffusion on a collection of rendered pose-annotated images \cite{deitke2023objaverse} excel at generating images of an object at new viewpoints. These models take an input image containing an object and the viewpoint change as input to generate diverse views.


\subsection{Dense-View Pose Estimation}
For estimating poses from image sets or continuous video streams,
traditional methods typically rely on identifying correspondences between hand-craft local features~\cite{lowe2004distinctive,liu2010sift,bay2006surf,tola2009daisy}, which then underwent validation and refinement with RANSAC and bundle adjustment. 
Recently, researchers have attempted to obtain correspondences via deep learning techniques, greatly improving the performance~\cite{choy2016universal,sarlin2020superglue,sun2021loftr}.
Some works~\cite{bundlesdfwen2023,wen2021bundletrack,sun2022onepose} effectively leverage correspondences to estimate object poses from the input video.

Template matching is another major approach.
Methods in this category aim to infer object pose with a set of dense reference images with known poses, 
leveraging category-specific priors~\cite{usman2022metapose,OCSKB} or category-agnostic priors~\cite{li2017deepim,zhao2022fusing,liu2022gen6d,he2022oneposeplusplus,ausserlechner2023zs6d,10350449_PoseMatcher,zhao2023locposenet,wang2024object_pose_via_aggregation}.
In particular, advancements in category-agnostic object pose estimation accuracy have been achieved through Gen6D~\cite{liu2022gen6d} and LocPoseNet~\cite{zhao2023locposenet}.
Wang \etal.~\cite{wang2024object_pose_via_aggregation} propose to extract features from a pre-trained diffusion model~\cite{rombach2022high} for the input image and dense reference images and then compute their similarity to determine the pose.
Additionally, the co-estimation of pose and 3D reconstruction has become feasible~\cite{wang2023pf,10350892_NeRF_Pose,yenchen2021inerf,park2019latentfusion}.
All these methods necessitate a rich set of input views to acquire sufficient information. 

\subsection{Sparse-View Pose Estimation}
\paragraph{Feature matching-based methods} \quad
Methods based on feature matching first identify pixel-level correspondences between views and then solve the fundamental matrix to get the relative pose. The correspondences can be obtained from hand-crafted features~\cite{liu2010sift,bay2006surf}, or learning-based features~\cite{sun2021loftr,jiang2021cotr,sarlin2020superglue}.
However, for these methods, a sufficient amount of available and effective features is a crucial prerequisite. Therefore, these approaches will encounter difficulties when images are sparsely captured with significant viewpoint changes.

\paragraph{Direct regression} \quad
Traditional methods to address the pose estimation problem with a few references are to constrain the problem within a limited and seen category~\cite{chang2015shapenet,mehta2017vnect,kanazawa2018end,lin2022learning_to,Castro_2023_WACV_CRT6D}.
Recently, some other works have started to explore the pose estimation of objects in unseen categories and proposed category-agnostic methods. However, many of them either rely on extra priors, such as 3D models~\cite{Cai_2022_CVPR,9021977_CorNet,9423117_3DPoseLite,grabner2020geometric,okorn2021zephyr,9880271_OSOP,Xiao2019PoseFromShape,nguyen2022template,hai2023pseudo}, and depth maps~\cite{GCCN,irshad2022centersnap,irshad2022shapo,he2022fs6d,ZSC,Li_2023_WACV_SDPose_needDepth,DTF_Net}.  
Melekhov \etal.~\cite{melekhov2017relative} and Rockwell \etal.~\cite{rockwell20228} have attempted to directly regress the relative pose between a single reference and a query image but struggled to achieve satisfactory accuracy.
E. Arnold \etal.~\cite{arnold2022map} proposed to use a single reference image to estimate relative pose between scenes to enable relocalization, which closely resembles our single reference setting. 
Recent works, such as RelPose~\cite{relpose2022}, and transformer-based approaches, Relpose++\cite{relpose++2023}, have further enhanced the performance of sparse view pose estimation. 
NOPE~\cite{nguyen2023nope} achieved similar objectives by predicting discriminative embeddings with a UNet, while GigaPose~\cite{nguyen2023gigapose} focused on estimating object poses using one reference image, primarily on CAD-like datasets. 
DUSt3R~\cite{wang2024dust3r} is a method that regresses pointmaps from input images, from which the relative pose can be determined using RANSAC with PnP.
Zhang \etal.~\cite{zhang2024camerasAsRays_raydiffusion} propose to estimate camera poses in a ray bundle representation.
However, many of the aforementioned works have exhibited limited generalization ability.

\paragraph{Pose estimation utilizing diffusion model} \quad
The most recent work, E2VG~\cite{e2vg} and IDPose~\cite{idpose2023}, belongs to this category. E2VG~\cite{e2vg} employed a diffusion-based generative model~\cite{liu2023zero123} to produce a series of novel RGB views conditioned on a reference image, which are then matched with the query image. IDPose~\cite{idpose2023} reversed the denoising process of the pre-trained diffusion model~\cite{liu2023zero123} to compute the relative pose.
These methods benefit from the generalization capability of the pre-trained generative model, Zero-1-to-3~\cite{liu2023zero123}, demonstrating the ability to bridge domain gaps and generalize across synthetic and real datasets. 
Inspired by E2VG~\cite{e2vg} and IDPose~\cite{idpose2023}, we further explore to effectively leverage the object prior encoded in the diffusion-based generative model to accurately estimate object pose. 
However, taking advantage of diffusion models may encounter challenges in scenarios when there is a large viewpoint change between two input images, resulting in poor matching results.
The proposed two-side matching module greatly alleviates this issue, demonstrating effectiveness in both real and synthesized datasets. 



\section{Method}
\label{sec:method}
\begin{figure}
    \centering
    \includegraphics[width=0.48\textwidth ]{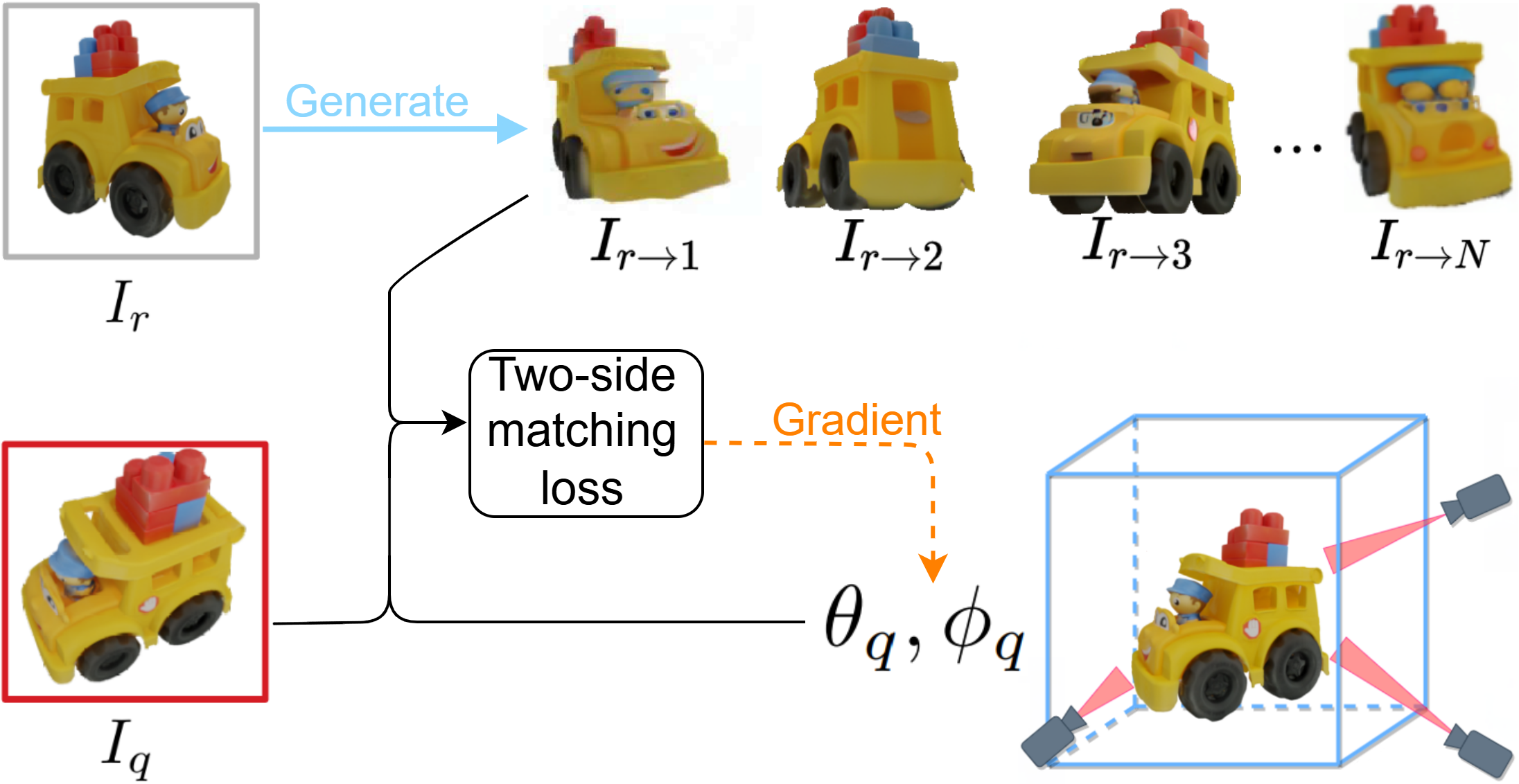}
    \caption{
The overview of our method. Given a reference image $I_r$ containing an object, our method is able to estimate the object pose of a query image $I_q$ containing the same object. We firstly utilize a pre-trained diffusion model to generate novel-view images. Then we estimate the elevation $\theta_q$ and azimuth $\phi_q$ of $I_q$ by minimizing the two-side matching loss, which will be explained in Sec~\ref{sec:two_side_matching}. 
} \label{fig:overview}
\end{figure}

Given an input reference image $I_r$ containing an object as shown in Fig.~\ref{fig:overview}, we assume that the object is located at the origin in the object coordinate system, and the object pose of the reference image is known, which is looking at the object with the up direction aligned with the gravity direction. Meanwhile, the viewpoint of $I_r$ has an elevation $\theta_r$ and an azimuth $\phi_r$. Then, given a query image $I_q$ containing the same object as shown in Fig.~\ref{fig:overview}, our target is to estimate the object pose of the query image $I_q$. We detect the object in $I_q$ and follow \cite{e2vg} to transform $I_q$ to look at the object. Thereafter, our target is transformed into estimating the elevation $\theta_q$ and azimuth $\phi_q$ of $I_q$ in the object coordinate system. 
We propose to use a diffusion generative model to solve this challenging problem by predicting the $\Delta \theta_{rq}$ elevation and $\Delta \phi_{rq}$ azimuth differences between the query image and the reference image.
Unlike previous object pose estimators~\cite{liu2022gen6d,relpose2022,relpose++2023}, our method does not necessitate training on the object and only relies on a single reference image for object pose estimation.

\subsection{Preliminaries: Diffusion Generative Models}

Diffusion generative models~\cite{ho2020denoising} are able to generate high-quality data. A diffusion model consists of a forward process and a reverse process.
The forward process gradually adds noise to the input image $I$ by
\begin{equation}
    I_t=\alpha_t I+ \sigma_t \epsilon,
    \label{eq:forward}
\end{equation}
where $t$ is an integer of the time step, $\alpha_t$ and $\sigma_t$ are predefined constants, and $\epsilon$ is a standard Gaussian noise. Then, the diffusion model learns the corresponding reverse process to gradually denoise the Gaussian noise step-by-step to recover the structure and generate the image. We denote the denoiser of the diffusion model as $\epsilon_\Theta(I_t; t)$, which learns to predict the noise $\epsilon$ from the noisy version $I_t$. The denoiser is iteratively applied to a Gaussian noise to generate an image.

Zero123~\cite{liu2023zero123} is a diffusion generative model to generate a novel-view image at a target viewpoint given an input view $I_{r}$, an elevation change $\Delta \theta$, an azimuth change $\Delta \phi$, and a radius change $\Delta r$. Thus, the denoiser of Zero123 is represented by $\epsilon_\Theta(I_t; t, I_r, \Delta \theta, \Delta \phi, \Delta r)$.

After being trained, at a diffusion time step $t$, the denoiser $\epsilon_\Theta$ can predict the noise of a noisy image. Here, we utilize the denoiser to estimate the $\Delta \theta_{rq}=\theta_q - \theta_r$ elevation and $\Delta \phi_{rq}=\phi_q - \phi_r$ azimuth differences between $I_q$ and $I_r$, so we omit the $\Delta r$ in the denoiser by setting $\Delta r=0$. This enables us to estimate the elevation and the azimuth of the query image.

\subsection{Naive Generation and Matching}
\label{sec:naive}

A straightforward solution is to find the $\Delta \theta_{rq}$ elevation and $\Delta \phi_{rq}$ azimuth differences between $I_q$ and $I_r$ by minimizing
\begin{equation}
    \mathop{\arg\min}_{\Delta \phi_{rq},\Delta \theta_{rq}} \mathbb{E}_{z} \|I_q - \mathbb{G}(z, I_r, \Delta \theta_{rq}, \Delta \phi_{rq})\|^2_2  \;,
    \label{eq:naive-target}
\end{equation}
where $z$ means a random variable representing the randomness in the generation, and $\mathbb{G}(z, I_r, \Delta \theta_{rq}, \Delta \phi_{rq})$ means the generated image from $I_r$ using $\Delta \theta_{rq}$ and $\Delta \phi_{rq}$. Problem~\ref{eq:naive-target} means that we want to find an azimuth difference and an elevation difference that can make the generated images look the most similar to the query image $I_q$. 

To solve the above problem, one solution is to use the denoising loss~\cite{poole2022dreamfusion} for optimization, which is adopted by IDPose~\cite{idpose2023}.
Another solution is to generate an image set 
$\mathcal{G}_r$ 
on novel viewpoints conditioned on $I_r$. Then, we match $I_q$ with the generated image set $\mathcal{G}_r$ to estimate $\Delta \phi_{rq}$ and $\Delta \theta_{rq}$, which is adopted by E2VG~\cite{e2vg}. 

\textit{Discussions} This naive generation and matching scheme does not perform well for this object-pose estimation problem. The main reason is that the quality of generation tends to drop as the viewpoint change increases. As shown in Fig.~\ref{fig:good_and_bad_match} (a), if $\Delta \theta_{rq}$ and $\Delta \phi_{rq}$ are large, it becomes challenging to match the generated image with $I_q$ due to the poor generation quality. Alternatively, we will estimate $\Delta \theta_{rq}$ and $\Delta \phi_{rq}$ by a new two-side generation matching scheme as stated in the following.

\subsection{Two-side Generation Matching}
\label{sec:two_side_matching}
\begin{figure} 
    \centering
    \includegraphics[width=.99\linewidth ]{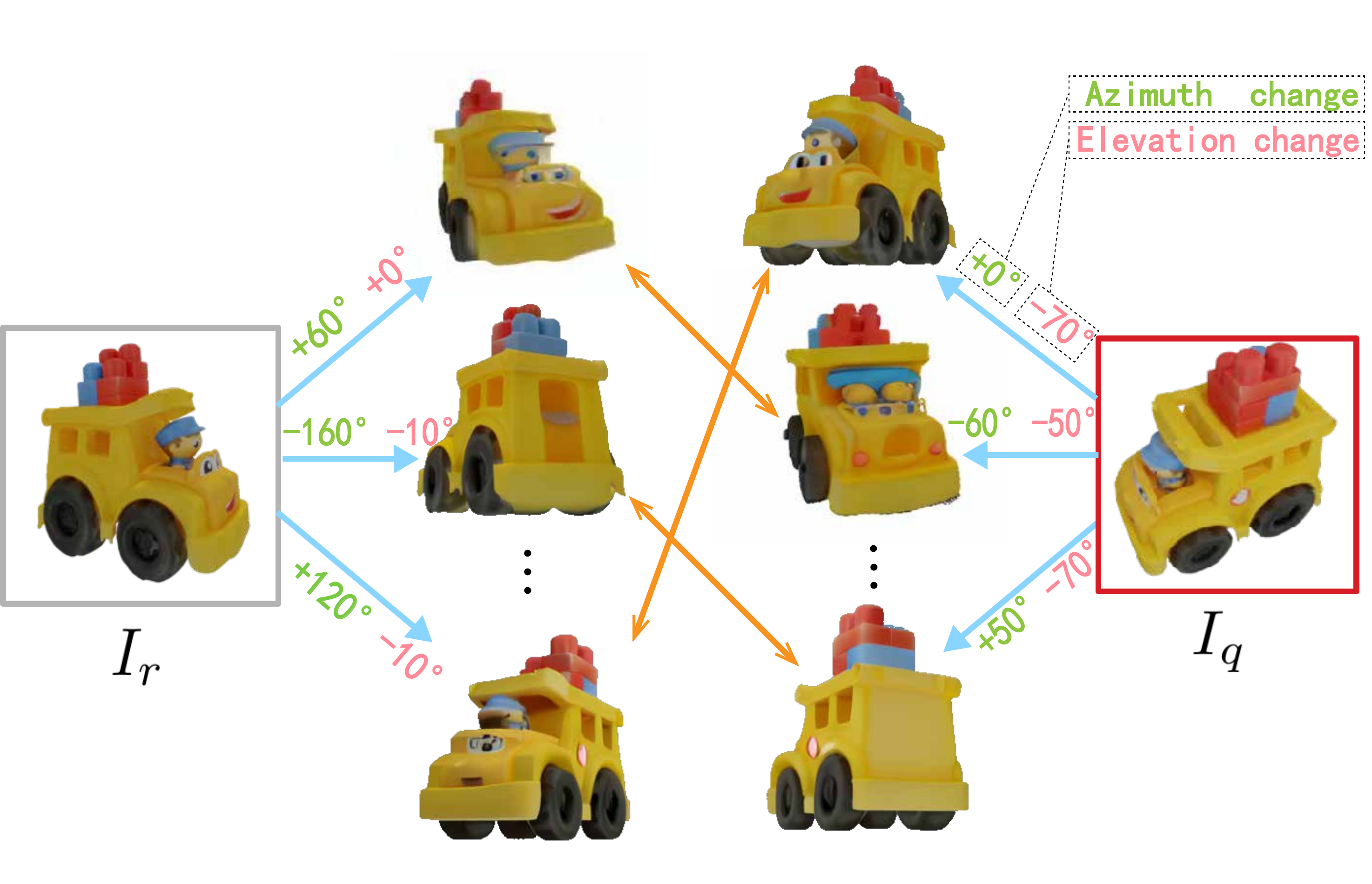}
    \vspace{-1.0em}
    \caption{ 
We generate images $\mathbb{G}(z, I_r, \Delta \theta_{ri}, \Delta \phi_{ri})$ of $I_r$ and $\mathbb{G}(z, I_q, \Delta \theta_{qi}, \Delta \phi_{qi})$ of $I_q$ on $N$ intermediate viewpoints $\{\phi_i,\theta_i | i = 1,...,N\}$, which are sampled evenly from the upper hemisphere of the object.  When the assumed $\phi_q,\theta_q$ is correct, for each $i$, $\mathbb{G}(z, I_q, \Delta \theta_{qi}, \Delta \phi_{qi}) $ and $ \mathbb{G}(z, I_r, \Delta \theta_{ri}, \Delta \phi_{ri})$ are well-matched.
    } \label{fig:set_to_set_match}
\end{figure}

For a better performance under large viewpoint changes, we resort to matching using generated images on the intermediate viewpoints between $I_q$ and $I_r$. 
We denote the image of the intermediate viewpoint as $I_i$
which has a predefined elevation $\theta_i$ and a predefined azimuth $\phi_i$. Here, we define a set of generated images on intermediate viewpoints as $\{I_i|i=1,...,N\}$ as shown in Fig.~\ref{fig:set_to_set_match}.
Then, we want to solve for $\phi_q$ and $\theta_q$ by
{\small
\begin{equation}
    \mathop{\arg\min}_{\theta_{q},\phi_{q}} \mathbb{E}_{z} \sum_{i=1}^{N} \|\mathbb{G}(z, I_q, \Delta \theta_{qi}, \Delta \phi_{qi}) - \mathbb{G}(z, I_r, \Delta \theta_{ri}, \Delta \phi_{ri})\|^2_2,
    \label{eq:set2set_argmin_Ez}
\end{equation}
}
where the $N$ intermediate viewpoints $\{\phi_i,\theta_i | i = 1,...,N\}$ are sampled evenly from the upper hemisphere of the object using the Fibonacci sphere algorithm~\cite{fibo_gonzalez2010measurement}  and the viewpoint changes are defined as $(\Delta \phi_{ri},\Delta \theta_{ri})=(\phi_i-\phi_r,\theta_i-\theta_r)$ and $(\Delta \phi_{qi},\Delta \theta_{qi})=(\phi_i-\phi_q,\theta_i-\theta_q)$ given the $(\phi_r, \theta_r)$. 
Here, we already know all the predefined viewpoints $\{(\phi_i,\theta_i)|i=1,2,...,N\}$ and $I_r$ with a viewpoint of ($\phi_r$, $\theta_r$). The only unknown here is the viewpoint $(\phi_q, \theta_q)$. Eq.~(\ref{eq:set2set_argmin_Ez}) means that we want to match the generated images $\mathbb{G}(z, I_r, \Delta \theta_{ri}, \Delta \phi_{ri})$ of $I_r$ and the generated images $\mathbb{G}(z, I_q, \Delta \theta_{qi}, \Delta \phi_{qi})$ of $I_q$ on these intermediate viewpoints to determine $(\phi_q, \theta_q)$ as shown in Fig.~\ref{fig:set_to_set_match}.

\vspace{-1.0em}
\paragraph{Score function} 
Problem~\ref{eq:set2set_argmin_Ez} is computationally intractable to solve because it requires matching all the generated images on these intermediate viewpoints and searching for an optimal viewpoint $(\phi_q,\theta_q)$.
Instead of directly solving it, we follow DreamFusion~\cite{poole2022dreamfusion} to approximate this problem by the loss in DreamFusion as
{\small
\begin{equation}
    \mathop{\arg\min}_{\theta_{q},\phi_{q}} \mathbb{E}_{t, \epsilon}\left\{  \sum_{i=1}^{N}    || \epsilon_\Theta(I_{r \to i,t}|I_q,\Delta \theta_{qi}, \Delta \phi_{qi})-\epsilon\ ||_2^2  \right\}.
    \label{eq:set2set_argmin_SDS}
\end{equation}
}
Here, we denote the intermediate images generated from $I_r$ as $I_{r\to i}$. We first generate images 
{\scriptsize
\[\mathcal{G}_r=\{I_{r\to i} | i=1,2,...,N\} = \{\mathbb{G}(z, I_r, \Delta \theta_{ri}, \Delta \phi_{ri}) | i=1,2,...,N\}\]
}
from $I_r$ on the intermediate viewpoints. Then, we add noise to $I_{r\to i}$ to obtain $ I_{r\to i,t}$ as Eq.~(\ref{eq:forward}). Successively, we feed $I_{r\to i,t}$ to the denoiser $\epsilon_\Theta$ of Zero123 to get the predicted noise $\epsilon_\Theta(I_{r\to i,t}|I_q,\Delta \theta_{qi}, \Delta \phi_{qi})$ from the query image $I_q$. Finally, we compute the matching distance by the L2 loss between the added noise and the predicted noise.

Compared to Problem~\ref{eq:set2set_argmin_Ez}, the problem in Eq.~(\ref{eq:set2set_argmin_SDS}) approximates the distances between the generated images by the score distillation loss.
Solving Problem~\ref{eq:set2set_argmin_SDS} only requires predicting the noise added to $I_{r \to i,t}$ conditioned on $(I_q,\Delta \theta_{qi}, \Delta \phi_{qi})$ and use the residuals in the noise prediction as the distance. However, there is still an expectation here in Problem~\ref{eq:set2set_argmin_SDS}, which needs to compute the residuals on all the time steps $t$ and all the Gaussian noises.

To simplify this, we fix the time step $t$ and compute an unbiased Monte Carlo estimate of the expectation by sampling $\epsilon$ for $M$ times. Empirically, we found that a fixed value for $t$ yields a good result. Thus, Problem~\ref{eq:set2set_argmin_SDS} now turns into
{\footnotesize
\begin{equation}
f( \theta_q,\phi_q ) = \frac{1}{M} \sum_{j}^{M} \left\{  \sum_{i=1}^{N}    || \epsilon_\Theta(I_{r \to i,t}^{(j)}|I_q,\Delta \theta_{qi}, \Delta \phi_{qi})-\epsilon^{(j)}\ ||_2^2  \right\},
\label{eq:score}
\end{equation}
}
where $\epsilon^{(j)}$ means the $j$-th sampled Gaussian noise. 
In the following, we call Eq.~(\ref{eq:score}) a score function and solve for $(\phi_q,\theta_q)$ by minimizing the score loss. To minimize it, we follow a coarse-to-fine search scheme to find the most plausible pose $(\phi_q,\theta_q)$.

\begin{figure}
    \centering
    \includegraphics[width=.99\linewidth]{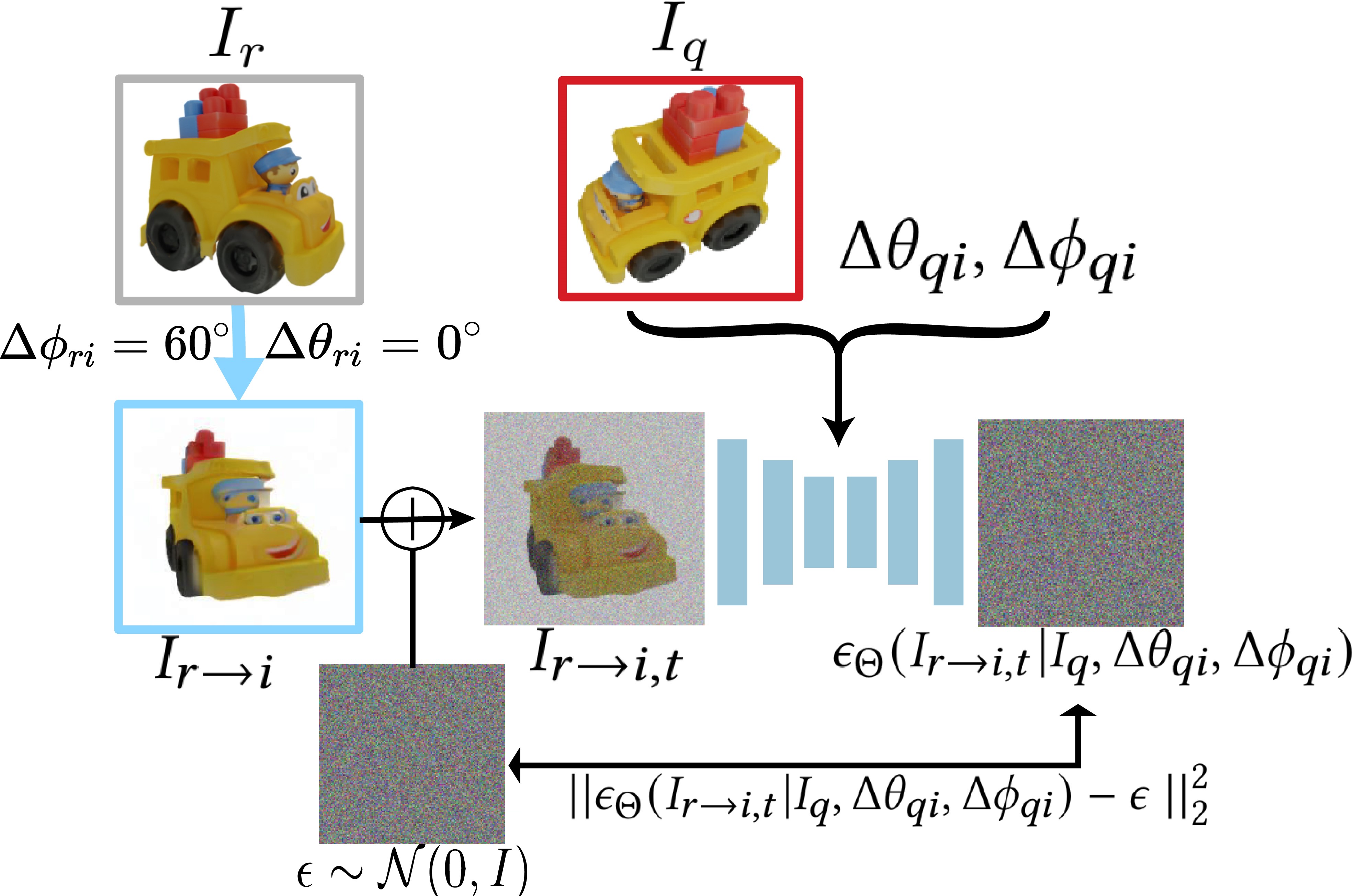}
    \caption{ As shown in Problem~\ref{eq:set2set_argmin_SDS}, we approximate the original two-side matching problem by minimizing the loss function proposed by Poole \etal.~\cite{poole2022dreamfusion}. For a generated image set $\mathcal{G}_r=\{I_{r\to i} | i=1,2,...,N\} = \{\mathbb{G}(z, I_r, \Delta \theta_{ri}, \Delta \phi_{ri}) | i=1,2,...,N\}$, we add noise to $I_{r\to i}$ to obtain $ I_{r\to i,t}$ as Eq.~(\ref{eq:forward}). 
Then, we employ the denoiser $\epsilon_\Theta$ of Zero123 to get the predicted noise $\epsilon_\Theta(I_{r\to i,t}|I_q,\Delta \theta_{qi}, \Delta \phi_{qi})$ 
and use the
$L_2$ loss
as the distance between the generated images.
} \label{fig:calculate_f}
\end{figure}

\vspace{-1.0em}
\paragraph{Grid-based coarse searching}  
In the coarse searching, we enumerate a set of predefined viewpoint candidates $(\theta_q,\phi_q)$ to compute the score function $f$ and select the one with the lowest score as the viewpoint. To reduce the search space, we first predict an initial elevation $\theta_q$ by the elevation predictor in One2-3-45~\cite{liu2023one}, which is also adopted in \cite{idpose2023,e2vg}. 
Then, we select a set of candidate azimuths $\phi_q$ ranging from $0^{\circ}$ to $360^{\circ}$ with an interval of $45^{\circ}$. Subsequently, we perform a search at $(\phi_q-30^{\circ},\phi_q-15^{\circ},\phi_q+15^{\circ},\phi_q+30^{\circ})$, and $(\phi_q-10^{\circ},\phi_q-5^{\circ},\phi_q+5^{\circ},\phi_q+10^{\circ})$ in the next round. Then, we assign the azimuth with the lowest matching score in this round to $\phi_q$. In the subsequent round, we use an interval $10^{\circ}$ around the initial elevation to calculate $f$ for $(\theta_q-20^{\circ},\theta_q-10^{\circ},\theta_q+10^{\circ},\theta_q+20^{\circ})$. This leads to a better $\theta_q$.

\vspace{-1.0em}
\paragraph{Refine coarse $(\theta_q,\phi_q)$ with gradient descent} Since the coarse grid-searching only enables a very coarse pose estimation with an interval of $10^\circ$, we will refine the pose by using gradient descent for optimization. Here, we randomly sample a noise $\epsilon$ to evaluate $f$ in Eq.~(\ref{eq:score}) and then compute the gradient $\frac{\partial f}{\partial \theta_q}$ and $\frac{\partial f}{\partial \phi_q}$ to update the viewpoint $(\theta_q,\phi_q)$. Here, we update the poses by SGD optimizer for a predefined number of steps.


\section{Experiments}

\subsection{Implementation Details}

\paragraph{Setups} 
The number of intermediate viewpoints $N$ is set to 64 as default. The number of sampling $M$ in Eq.~(\ref{eq:score}) is set to 4 for both the searching stage and the refinement stage. We adopt a fixed value of time step $t=0.4$ (with 1000 denoising steps in total).
The refinement process is repeated 3 times by default.  

\vspace{-1.0em}
\paragraph{Baselines}
To assess the effectiveness of our proposed framework, we compare our method against three kinds of baseline methods. The first category is the matching-based pose estimation methods, including LoFTR~\cite{sun2021loftr} and SIFT~\cite{lowe2004distinctive}. Matching-based methods first build dense~\cite{sun2021loftr} or sparse~\cite{lowe2004distinctive} correspondences and then solve the relative pose from correspondences by a RANSAC algorithm. For the SIFT-based method, we also consider incorporating single-view depth estimation methods like Zoe-Depth~\cite{bhat2023zoedepth} to solve the pose directly by a 2D-to-3D PnP algorithm or a 3D-to-3D Procrustes algorithm. The second category includes regression-based methods, which are Relative Pose Regression (RPR)~\cite{arnold2022map}, 3DAHV~\cite{3DAHV}, and RelPose++~\cite{relpose++2023}. 
RPR~\cite{arnold2022map} applies CNNs to directly regress the relative poses from the reference view to the query view.  
3DAHV~\cite{3DAHV} introduces a 3D-aware verification module that explicitly applies 3D transformations to the 3D object representations learned from two input images. It evaluates multiple pose hypotheses and selects the most reliable one as the predicted pose.
RelPose++~\cite{relpose++2023} is an improved version of RelPose~\cite{relpose2022}, which directly uses Transformers to regress the poses for objects. Note that we only adopt the two-view setting of RelPose++ here. The third category includes diffusion-based pose estimation methods, E2VG~\cite{e2vg} and IDPose~\cite{idpose2023}. Both diffusion-based methods generate novel-view images by Zero123~\cite{liu2023zero123} to estimate the relative poses between images with large viewpoint changes, which is similar to our method. However, as discussed in Sec.~\ref{sec:naive}, both methods only consider directly generating the image on the query viewpoint, while our methods will match the images generated on the intermediate viewpoints for both query and reference images.

For SIFT-ZoeDepth-based methods and Relative Pose Regression we adopt the implementation from \cite{arnold2022map}. For other methods, we adopt the implementation provided in their official GitHub repositories.

\vspace{-1.0em}
\paragraph{Datasets} \quad 
We conducted an evaluation using the same benchmark as E2VG~\cite{e2vg}, where the experiments were performed on the GSO~\cite{gso2022google} and NAVI~\cite{jampani2023navi} datasets. 
The GSO dataset~\cite{e2vg} is a synthetic object dataset that consists of approximately 1,000 3D-scanned objects. The GSO~\cite{gso2022google} testing set contains 23 objects. For each object, E2VG renders multiple images with evenly sampled viewpoints on the upper hemisphere of the object and randomly samples 1 reference view and 20 query views.
The NAVI dataset~\cite{jampani2023navi} is a real object dataset that contains images of the same objects captured in various environments and viewpoints. The NAVI dataset provides high-quality intrinsic and extrinsic matrices. There are 36 objects in NAVI dataset in total and E2VG filters out some inappropriate objects such as symmetric objects and remains 27 objects as the testing set. Each object contains 20 reference-query pairs for evaluation, except for those objects with fewer than 20 images, in which case we use all available images.

\vspace{-1.0em}
\paragraph{Metrics} \quad 
To evaluate baseline methods and our method, we follow previous works~\cite{sun2021loftr,relpose++2023,relpose2022,e2vg} to use the accuracy under a specific degree as the metrics. For the rotation accuracy, we compute the relative rotation between the ground-truth rotation $\mathbf{R}_{\text{gt}}$ and the predicted rotation $\mathbf{R}_{\text{pr}}$ by $\mathbf{R}_{\text{gt}}^{\intercal} \mathbf{R}_{\text{pr}}$ and then transform this rotation into the axis-angle form. The resulting rotation angle of this rotation matrix is regarded as the rotation error in angle. 
We report the rotation accuracy under 15$^\circ$ and 30$^\circ$.

\subsection{Comparisons with Baseline Methods}

\begin{table}
\caption{The quantitative comparison results on the synthesized dataset GSO~\cite{gso2022google}, asd the real dataset NAVI~\cite{jampani2023navi}. We report the proportion of angular errors within 15$^\circ$ and 30$^\circ$. 
}
\vspace{-0.5em}
\label{table:compareALL}
\centering
\resizebox{.49\textwidth}{!}{
\begin{tabular}{lcccccccc}
\toprule
\multirow{3}{*}{Methods}                  & \multicolumn{2}{c}{NAVI~\cite{jampani2023navi}}  & \multicolumn{2}{c}{GSO~\cite{gso2022google}} \\
 &
  \multicolumn{2}{c}{Rotation  Accuracy} &
  \multicolumn{2}{c}{Rotation Accuracy} \\

 &
  \multicolumn{1}{c}{15$^\circ$} &
  \multicolumn{1}{c}{30$^\circ$} &
  \multicolumn{1}{c}{15$^\circ$} &
  \multicolumn{1}{c}{30$^\circ$} \\ 
\midrule
SIFT~\cite{lowe2004distinctive}+ZoeDepth~\cite{bhat2023zoedepth}+PnP   & 19.66 & 25.55          & 7.17 & 13.04        \\
SIFT~\cite{lowe2004distinctive}+ZoeDepth~\cite{bhat2023zoedepth}+Procrustes    & 16.65 & 26.72           & 5.00 & 14.78       \\
RPR~\cite{arnold2022map}    & 17.23 & 33.99            & 4.57 & 15.87   \\
LoFTR~\cite{sun2021loftr}      & 16.59 & 27.99          &  20.65 & 29.57      \\
3DAHV~\cite{3DAHV} & 28.23 & 48.34              & 16.74 & 38.91    \\
IDPose~\cite{idpose2023} & 10.09 & 36.66                   & 20.43 & 40.43     \\
Relpose++~\cite{relpose++2023}       &  24.33 &  40.05            & 15.65 &  32.17       \\
E2VG(N=64)~\cite{e2vg} & 42.69 & 64.21      & 39.30 & 55.86    \\
E2VG(N=128)~\cite{e2vg}  & 43.16 & 66.47    & 40.43 & 57.61   \\
\textbf{Ours}   & \textbf{55.32} & \textbf{82.14}          & \textbf{58.26} & \textbf{72.61}     \\
\bottomrule
\end{tabular}}
\end{table}

\begin{table}[ht]
\caption{The quantitative comparison results under large viewpoint changes $\delta \ge 120^{\circ}$ and $150^{\circ}$ respectively. We report the proportion of angular errors within 15$^\circ$ and 30$^\circ$ on GSO dataset.
}
\vspace{-0.5em}
\label{table:compareALL_largeViewpointChange}
    \centering
    \small
        \begin{tabular}{lcccc}
        \hline
        \multicolumn{1}{c}{\multirow{3}{*}{Methods}} & \multicolumn{2}{c}{$\delta \ge 120^{\circ}$} & \multicolumn{2}{c}{$\delta \ge 150^{\circ}$} \\
         & \multicolumn{2}{c}{Rotation Accuracy } & \multicolumn{2}{c}{Rotation Accuracy } \\

        \multicolumn{1}{c}{}                   & $15^{\circ}$         & $30^{\circ}$         & $15^{\circ}$         & $30^{\circ}$         \\ \hline
        Relpose++\cite{relpose++2023}       &  4.07 &  17.89         & 3.64 &  16.62    \\
        3DAHV\cite{3DAHV} & 13.82 & 21.38             & 14.55 &  20.00   \\
        IDPose\cite{idpose2023} & 6.05 & 21.14                 & 9.09 &  20.00    \\
        E2VG\cite{e2vg} & 19.51 & 34.14       & 16.36 &  32.73  \\
        \textbf{Ours}   & \textbf{31.71} & \textbf{49.59}          & \textbf{21.82} & \textbf{47.27}     \\ \hline
        \end{tabular}
\end{table}

We report the quantitative comparison results in Table.~\ref{table:compareALL}. It can be seen that our method achieves the best performance and outperform baselines by a large margin. 

This demonstrates the effectiveness of using the intermediate viewpoints in the generation and matching, which enables us to find an accurate object pose.
Since the main difference between our pipeline, E2VG, and ID-Pose lies in the matching scheme, where we adopt two-side matching while E2VG and IDPose use naive matching, the comparison between our method and them serves as an ablation study of the two-sided matching.

To highlight our superior advantage under large viewpoint changes, we additionally conduct a statistical analysis for viewpoint changes of $\delta \ge 120^{\circ}$ and $150^{\circ}$, respectively.
The analysis is performed on the GSO dataset, which comprises images with viewpoints uniformly distributed across the upper hemisphere, thus providing sufficient cases with large pose changes.
The result is present in Table.~\ref{table:compareALL_largeViewpointChange}. It's clear that
our method achieves approximately 1.5 times the accuracy of the best baseline method, 
indicating that the proposed approach performs significantly better, particularly in scenarios involving large viewpoint changes.

\begin{figure*}
\begin{minipage}{0.99\textwidth}
    \begin{center}
    \begin{minipage}{0.99\textwidth}
    \includegraphics[width=0.99\linewidth]{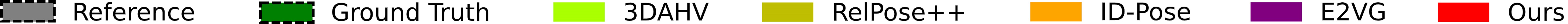}
    \end{minipage}
    \end{center}
    \vspace{-2.5em}
\end{minipage}
\begin{center}
    \begin{minipage}{0.99\textwidth}
        \vspace{+0.0em}
    \end{minipage}
    \begin{minipage}{0.98\textwidth}
    \begin{center}
        \begin{minipage}{0.24\textwidth}
        \includegraphics[width=0.99\linewidth]{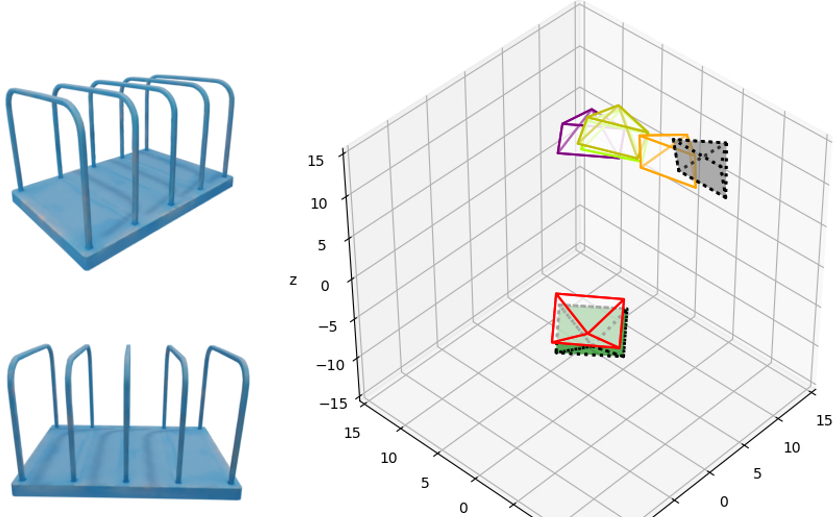}
        \end{minipage}
        \begin{minipage}{0.24\textwidth}
        \includegraphics[width=0.99\linewidth]{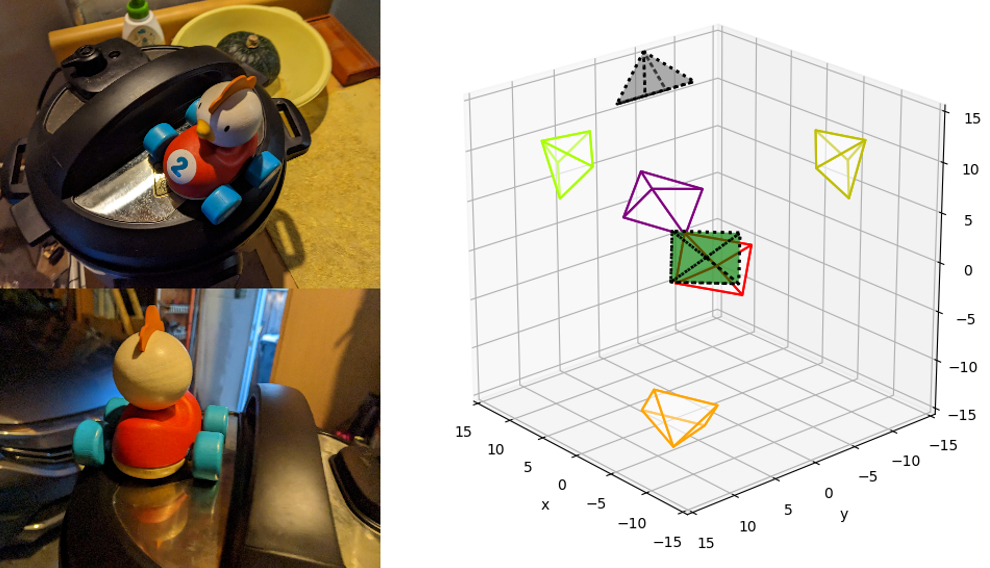}
        \end{minipage}
        \begin{minipage}{0.24\textwidth}
        \includegraphics[width=0.99\linewidth]{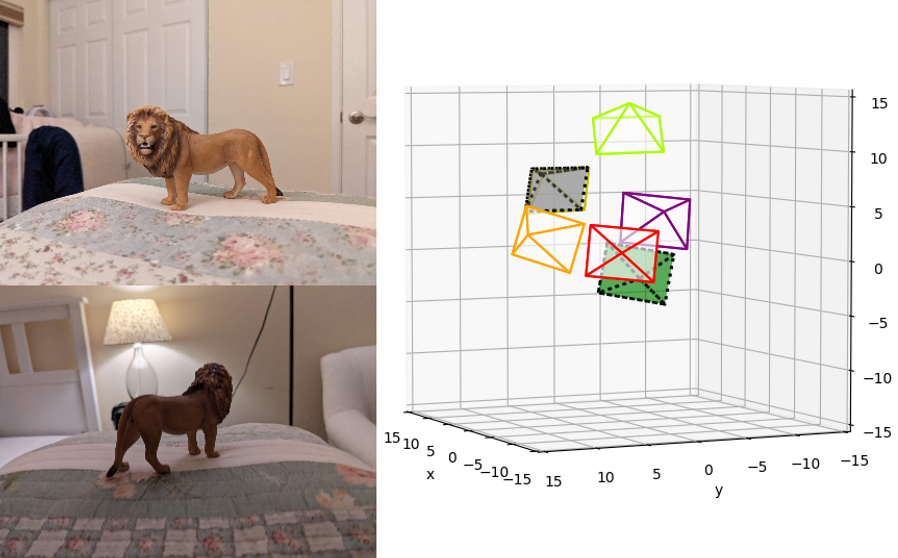}
        \end{minipage}
        \begin{minipage}{0.24\textwidth}
        \includegraphics[width=0.99\linewidth]{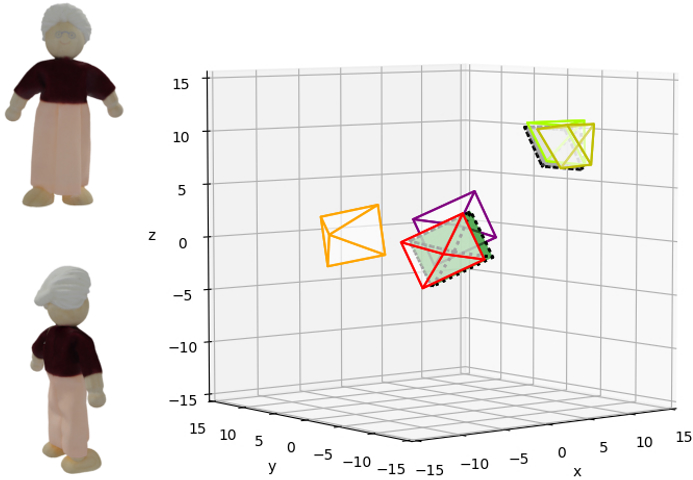}
        \end{minipage}
    \end{center}
    \end{minipage}
\end{center}
\vspace{-1.0em}
\caption{Visual Comparison on testing sets~\cite{gso2022google,jampani2023navi}. More results can be found in the supplementary materials. }
\label{fig:comp_testsets}
\end{figure*}

We demonstrate visual comparisons between our method and the baselines on GSO dataset~\cite{gso2022google} and NAVI dataset~\cite{jampani2023navi} in 
Fig.\ref{fig:comp_testsets}.
Consistent with the quantitative results, our method delivers the most visually satisfying pose estimation even under significant viewpoint changes. While the most of baseline methods struggle to find a plausible solution, our proposed method achieves a relative pose estimation closest to the ground truth. Although E2VG~\cite{e2vg} and IDPose~\cite{idpose2023} also capitalize on the advantages of diffusion generative models, our method outperforms them due to the more effective and efficient utilization of the provided diffusion image priors with the intermediate viewpoint images. More results on the NAVI dataset are shown in Fig.~\ref{fig:teaser}.

\subsection{Ablation Analysis}
In this section, we explain and discuss the effectiveness of different designs in our architecture, including the number of Monte-Carlo sampling $M$, the intermediate viewpoint number $N$ in the generation, the time step $t$ used in Eq.~(\ref{eq:score}) as well as the number of refinement iterations. 
We present results on NAVI~\cite{jampani2023navi} dataset in this section. More ablation analysis can be found in supplementary materials.


\begin{figure*}
    \centering
    \includegraphics[width=0.9\linewidth ]{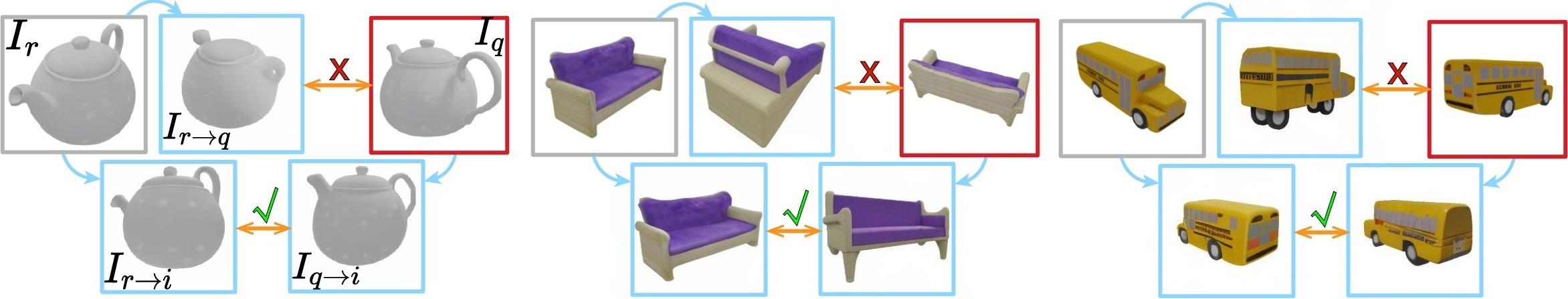}
    \vspace{-0.5em}
    \caption{ 
More examples of generated images on intermediate viewpoints, similar to Fig.~\ref{fig:good_and_bad_match}.
    } \label{fig:abla_match}
\end{figure*}

\vspace{-1.0em}
\paragraph{The number of the Monte-Carlo sampling $M$}\quad 
The effect of adopting different values of $M$ in the searching stage can be found in Table.~\ref{table:repeat_size}. 
The proposed method achieves improved performance with increased $M$ from 1 to 4, but such a trend becomes less noticeable when increasing from 4 to 8. Considering that when $M=8$, our method takes nearly twice as long as $M=4$, we set $M=4$ as the default configuration for the efficiency.

\vspace{-1.0em}
\paragraph{Intermediate viewpoint number $N$} \quad 
To assess how the number of intermediate viewpoints 
$N$ affects estimation accuracy, we evaluate our performance with different numbers of generated images. The results are presented in Table.~\ref{table:set_size}, indicating a notable enhancement in performance as $N$ increases from 16 to 64. Further increments in $N$ lead to marginal improvements in accuracy. Hence, to balance efficiency and effectiveness, we set 64 as the default.

\vspace{-1.0em}
\paragraph{Time step $t$ in Eq.~(\ref{eq:score})} \quad 
The impact of different values of $t$ is illustrated in Table.~\ref{table:timestep}.  Our approach achieves optimal performance when $t$ is configured to 0.4, suggesting that the method is most effective for a moderate level of noise.

\begin{figure} 
    \centering
    \includegraphics[width=.99\linewidth ]{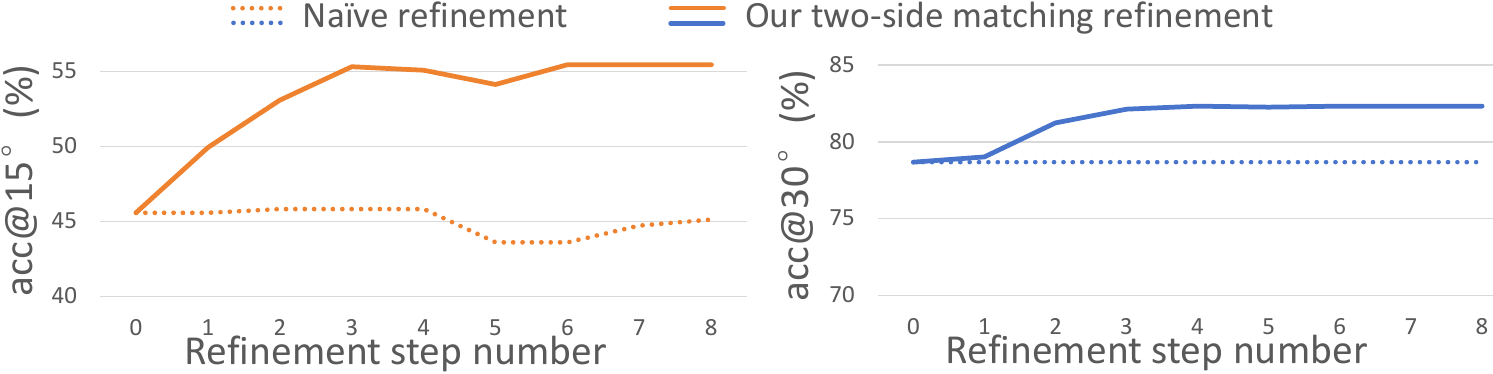}

    \vspace{-0.5em}
    \caption{Ablation studies on the refinement number and generating \& matching scheme on the NAVI dataset.} \label{fig:number_of_iter}
\end{figure}

\vspace{-1.0em}
\paragraph{Refinement step number and two-side generating \& matching } \quad 
We further investigated how the number of refinement iterations in the pose refinement process affects the quality of relative pose estimation, as shown by the solid line in Fig.~\ref{fig:number_of_iter}. We also present the result of the naive refinement approach, which directly generates and matches the image as discussed in Sec.~\ref{sec:naive}, represented by the dashed line in Fig.~\ref{fig:number_of_iter}. As we can see from the results, our two-side generating and matching scheme improves the accuracy at 15$^\circ$ while the naive generating and matching scheme fails to yield a significant improvement. As observed, our method process primarily benefits accuracy at 15$^\circ$, and the resulting accuracy is continuously improved with 0 to 3 iterations. Therefore, we choose to use 3 refinement iterations as a balance and stop using further refinement steps.

\vspace{-1.0em}
\paragraph{Generation quality on the intermediate viewpoint}\quad 
To further demonstrate our idea about using the intermediate viewpoints for matching, we show more examples of the generated images on the intermediate viewpoints in Fig.~\ref{fig:abla_match}.
As shown, the images generated from the query image and the reference images show strong similarity, which enables us to find a correct relative pose. On the contrary, the image directly generated from $I_r$ at the viewpoint of $I_q$ fails to match the query image, leading to incorrect poses.

\begin{table}
\caption{Ablation studies on the number of intermediate viewpoint number $N$ in the search stage on the NAVI dataset.}
\vspace{-0.5em}
\label{table:set_size}
\centering

\resizebox{.45\textwidth}{!}{
\begin{tabular}{l|llll} \hline 
\#Intermediate viewpoint $N$ &16& 32& 64& 128 
\\  \hline 

Rotation  Accuracy @ 15$^\circ$ & 38.67& 35.33& 45.58& 48.39   \\
Rotation  Accuracy @ 30$^\circ$ & 69.81& 73.14& 78.69& 79.37   \\ 
\hline\end{tabular}
}
\end{table}

\begin{table}
\caption{Ablation studies on the number of the Monte-Carlo sampling $M$ in the search stage on the NAVI dataset.}
\vspace{-0.5em}
\label{table:repeat_size}
\centering
\resizebox{.45\textwidth}{!}{
\begin{tabular}{l|llll} \hline 
\#Monte-Carlo sampling $M$ & 1	& 2	& 4	& 8 \\ \hline 
Rotation  Accuracy @ 15$^\circ$ & 44.16&  45.53& 45.58& 45.87  \\
Rotation  Accuracy @ 30$^\circ$ & 73.94& 78.64& 78.69& 79.93
 \\ \hline\end{tabular}}
\end{table}
\begin{table}
\caption{Ablation studies on the time step $t$ in the search stage on the NAVI dataset.}
\vspace{-0.5em}
\label{table:timestep}
\centering

\resizebox{.45\textwidth}{!}{
\begin{tabular}{l|llll}
\hline 
Time step $t$ & 0.2   & 0.4   & 0.6   & 0.8   \\ \hline 

Rotation  Accuracy @ 15$^\circ$ & 39.78 & 45.58 & 38.67 & 35.33 \\
Rotation  Accuracy @ 30$^\circ$ & 64.81 & 78.69 & 60.36 & 45.64 \\ 
\hline 
\end{tabular}
}
\end{table}

\subsection{Runtime Analysis}
Our runtime is approximately proportional to 
the sample number $M$ and the intermediate viewpoint number $N$. With the default configuration, to process a query view, our method takes 6.5 seconds to search for a coarse pose and 1.3 seconds to refine on a single A100 GPU. 

This processing time restricts the application in some real-time scenarios. To mitigate this limitation, we propose several acceleration techniques, including intermediate viewpoint pruning, feature reuse, etc, to accelerate the inference while maintaining comparable accuracy. We refer to the method that incorporates these acceleration techniques as the light version of our method. For this light version method, processing a query image takes 1.12 seconds, with only a slight reduction in $Racc@15$ and $Racc@30$ by 1\% and 4\%, respectively. This indicates that the accuracy of our method still significantly outperforms other baselines by a large margin. Details of the proposed acceleration techniques are provided in the supplementary material.

Finally, note that our method is easily parallelizable. Specifically, we can allocate each GPU to simultaneously compute different candidates, thereby reducing the total processing time approximately by a factor equal to the number of GPUs. With multiple GPUs, we can process each query image within one second.

\subsection{Limitations}
Firstly,
our method takes advantage of the diffusion models to provide object priors for pose estimation. Thus, our prediction ability is limited by the diffusion model and the performance of our method could be further improved with a better diffusion model like SV3D~\cite{voleti2024sv3d} or a multi-view diffusion model like SyncDreamer~\cite{liu2023syncdreamer}, Wonder3D~\cite{long2023wonder3d} and MVDiffusion++~\cite{tang2024mvdiffusion++}. 

Secondly, due to the difficulty of the two-view RGB setting, estimated poses are not precise to a few degrees.
Thirdly, as Zero123 is object-centric, our method cannot directly handle scenes. 
We show a straightforward way to extend our approach to scenes in Section 6 in the supplementary material.


\section{Conclusion}
In this paper, we propose a new object pose estimator that not only generalizes to unseen objects but also only requires one RGB reference image of the object. To achieve this target, we utilize the object prior encoded in a diffusion model to ensure the generalization ability. We find that a naive generation and matching scheme leads to poor performances especially when the viewpoint change is large. 
Thus, to improve the pose estimation accuracy, we propose a two-side generating and matching scheme to denoise the images from both reference image and query image and match the noise on the intermediate viewpoints.
Experiments on both real and synthetic datasets demonstrate the effectiveness of our design and greatly expand the potential of object pose estimation in 3D reconstruction, robot manipulation, AR, VR, simultaneous localization and mapping (SLAM), and autonomous driving.

{\small
\bibliographystyle{ieee_fullname}
\bibliography{egbib}
}

\end{document}